%% file: main.tex
\documentclass[fleqn,10pt]{wlscirep}

\input{Files/packages}
\input{Files/Title}
\input{Files/abstract}

\begin{document}
\flushbottom
\maketitle
\thispagestyle{empty}

\input{Sections/01_Intro}
\input{Sections/02_Methodology}

\input{Sections/02_Results}

\input{Sections/03_Discussion}

\bibliography{sample}

\section*{Author contributions statement}
A.R.E. and K.M.G. conceived the experiment(s),  A.R.E. and K.M.G. conducted the experiment(s), All authors analysed the results.  All authors reviewed the manuscript. 

\section*{Competing interests}
The authors declare no competing interests.

\section*{Funding}
This work received no fund from any organization.

\section*{Data Availability}
The datasets generated and/or analysed during the current study are available in the JAAD Website, \url{https://data.nvision2.eecs.yorku.ca/JAAD_dataset/}

\section*{Additional information}

\end{document}

%% file: Files/packages.tex
\usepackage[utf8]{inputenc}
\usepackage[T1]{fontenc}

\usepackage{cite}
\usepackage{algorithm} 
\usepackage{algorithmic}  
\usepackage[algo2e]{algorithm2e} 
\usepackage{amsmath,amssymb,amsfonts}
\usepackage{algorithm}
\usepackage{graphicx}
\usepackage{textcomp}
\usepackage{mwe} 
\usepackage{booktabs} 
\usepackage{siunitx} 
\usepackage{tabularx} 
\usepackage{lipsum} 
\usepackage{multirow} 
\def\BibTeX{{\rm B\kern-.05em{\sc i\kern-.025em b}\kern-.08em
    T\kern-.1667em\lower.7ex\hbox{E}\kern-.125emX}}

\usepackage{enumitem}
\usepackage{lineno}
\usepackage{lettrine}

\usepackage{url}
\usepackage{moreverb}

\usepackage{graphicx}
\usepackage{subcaption}
\usepackage{float}
\usepackage{amsmath}
\usepackage{array}
\usepackage{multirow}
\graphicspath{ {Figures/} }

\usepackage{scalerel}
\usepackage{tikz}
\usetikzlibrary{svg.path}

\definecolor{orcidlogocol}{HTML}{A6CE39}
\tikzset{
  orcidlogo/.pic={
    \fill[orcidlogocol] svg{M256,128c0,70.7-57.3,128-128,128C57.3,256,0,198.7,0,128C0,57.3,57.3,0,128,0C198.7,0,256,57.3,256,128z};
    \fill[white] svg{M86.3,186.2H70.9V79.1h15.4v48.4V186.2z}
                 svg{M108.9,79.1h41.6c39.6,0,57,28.3,57,53.6c0,27.5-21.5,53.6-56.8,53.6h-41.8V79.1z M124.3,172.4h24.5c34.9,0,42.9-26.5,42.9-39.7c0-21.5-13.7-39.7-43.7-39.7h-23.7V172.4z}
                 svg{M88.7,56.8c0,5.5-4.5,10.1-10.1,10.1c-5.6,0-10.1-4.6-10.1-10.1c0-5.6,4.5-10.1,10.1-10.1C84.2,46.7,88.7,51.3,88.7,56.8z};
  }
}

\newcommand\orcidicon[1]{\href{https://orcid.org/#1}{\mbox{\scalerel*{
\begin{tikzpicture}[yscale=-1,transform shape]
\pic{orcidlogo};
\end{tikzpicture}
}{|}}}}

\usepackage{hyperref} 

%% file: Files/Title.tex
\title{VIT-Ped: Visionary Intention Transformer for Pedestrian Behavior Analysis}

\author[1,2,+]{Aly R. Elkammar}
\author[1,2,+]{Karim M. Gamaleldin}
\author[1,2,*]{Catherine M. Elias}
\affil[1]{German University in Cairo, Faculty of Media Engineering and Technology, Computer Science and Engineering Department, Cairo, Egypt}
\affil[2]{C-DRiVeS Lab: Cognitive Driving Research in Vehicular Systems, Cairo, Egypt}

\affil[*]{catherine.elias@ieee.org}

\affil[+]{these authors contributed equally to this work}

%% file: Files/abstract.tex
\keywords{Autonomous Vehicles, Pedestrian Intention, Prediction, Transformer, ViVit, Perception}

\begin{abstract}
Pedestrian Intention prediction is one of the key technologies in the transition from level 3 to level 4 autonomous driving. To understand pedestrian crossing behaviour, several elements and features should be taken into consideration to make the roads of tomorrow safer for everybody. We introduce a transformer / video vision transformer based algorithm of different sizes which uses different data modalities .We evaluated our algorithms on popular pedestrian behaviour dataset, JAAD, and have reached SOTA performance and passed the SOTA in metrics like Accuracy, AUC and F1-score. The advantages brought by different model design choices are investigated via extensive ablation studies.
\end{abstract}

%% file: Sections/01_Intro.tex
\section*{Introduction and Related Work}
Over the last several years, numerous accidents have occurred in which 94\% are caused by human error. For example, 25\% of fatal accidents are caused by excessive speed. Such problems suggest our need for automating the driving activity. Therefore, it is expected that by the year 2045, 60\% of all cars in the US will be fully autonomous. However, exploring higher levels of automation in cars comes with its challenges. According to the World Health Organization (WHO), approximately 1.35 million deaths and 20-50 million injuries occur annually. Without the correct safety measures and algorithms, those numbers could skyrocket, making the urban environment more dangerous than ever. For this reason, the development of algorithms for different urban scenarios is a major focus of research and development. Besides obstacle detection and advanced driver-assistance systems (ADAS), pedestrian intention/behavior predictions are of vital importance, as they anticipate pedestrian movements, allowing for more timely and appropriate decisions, thus making the roads of tomorrow safer for everyone.

Over the last decade, pedestrian intention prediction has been a hot research topic, with different datasets, algorithms, and features being used to try and address such challenges. However, due to the fact how complex human beings are and how hard to predict their intention such a task is still researched heavily trying to achieve the best possible performance. In the literature, the terms intention, action, and behavior are often used interchangeably to describe pedestrian behavior. In our work, we try to anticipate the intention of the pedestrian which can be defined as the mental state that represents the pedestrians' commitment and planning which can't be observed by classical methods.

Throughout the years, many algorithms, modalities, and features have been proposed to tackle the task at hand. In the earlier days and due to the limitations of the GPUs' technologies, Support vector machines (SVMs) were the preferred models for such tasks. For example \cite{kotseruba2016joint} used a small Convolutional neural network to encode the visual data to feed it to an SVM model to classify the crossing and not crossing behavior. In more recent work, input sequences were used so we could capture both the spatial and the temporal dimensions, can exploit dependencies between different frames in the sequences. This approach is called \textbf{Spatio-temporal modeling}, which can be achieved by several steps. Firstly we need to extract or encode the spatial features using Feedforward neural networks (fc) for non-visual data and 2D Convolutional neural networks (CNNs) or graph convolutional neural networks for visual data. After that, we need to feed these features to an architecture that can capture the temporal dependencies like Recurrent Neural Networks (RNNs), long-short term memory (LSTMs) and Gated Recurrent units (GRUs). For example, \cite{kotseruba2020they} used a 2D CNN to encode the visual features then passed the encoded features to a RNN to encode the temporal features between different frames and lastly the encoded spatio-temporal features are fed into a fc layer for the action classification. 

Unlike the approach mentioned above, other approaches can be found in the literature trying to learn the spatial and temporal features simultaneously. For example \cite{kotseruba2021benchmark} and  \cite{tran2014learning} used 3D Convolutinoal Neural Networks, which captures the spatio-temporal features unlike their 2D counterpart. It is able to do this by changing the 2D kernels responsible for convolution and pooling operation by their 3D versions. Another approach that was used by \cite{shi2015convolutional} is to extend the LSTM adding a convolutional structure, called ConvLSTM, which can be used to learn spatial and temporal features effectively.

Recently, the introduction of transformers \cite{vaswani2017attention} in 2017 for natural language translation marked a significant development in the deep learning field as they outperformed other architectures not only in the NLP field but also with several modifications \cite{dosovitskiy2020image} \cite{arnab2021vivit} they also dominated the computer vision field. Such dominance was mainly because of the use of attention mechanisms which weigh the importance of different parts of the input data, allowing for parallel processing and capturing long-range dependencies within sequences with no loss of information. Lately, the use of transformer-based algorithms can be seen in the literature, where \cite{lorenzo2021capformer} , \cite{rasouli2023pedformer} and \cite{lorenzo2021intformer} tackled pedestrian intention prediction replacing the need for RNNs and/or CNNs. Others \cite{kotseruba2021benchmark} and \cite{azarmi2024pip}, tried to incorporate the attention mechanism with normal RNNs and CNNs to tackle the task at hand.

In this work, we propose a new method, PedViViT, which uses a variation of the state-of-the-art Video Vision Transformer model \cite{arnab2021vivit} for visual data and a transformer-based encoder \cite{vaswani2017attention} for non-visual data to predict the pedestrian's intention of whether he/she will cross or not. To summarize, our contributions in this paper are:
\begin{enumerate}
  \item A multi-modal model for pedestrian intention prediction. The model is based on a transformer architecture model, called Video Vision Transformer (ViVit) \cite{arnab2021vivit}.
  \item SOTA performance with a small model (x6 smaller than PCPA), which is more suitable for real time applications.
  \item SOTA performance with a non-visual model (x15 smaller than PCPA) which uses only numerical inputs such as bounding box coordinates
  \item Different features and modalities are explored in the ablation studies.
\end{enumerate}

%% file: Sections/02_Methodology.tex
\graphicspath{{Figures/}}
\begin{figure*}[t]
\centering
\includegraphics[width=\textwidth]{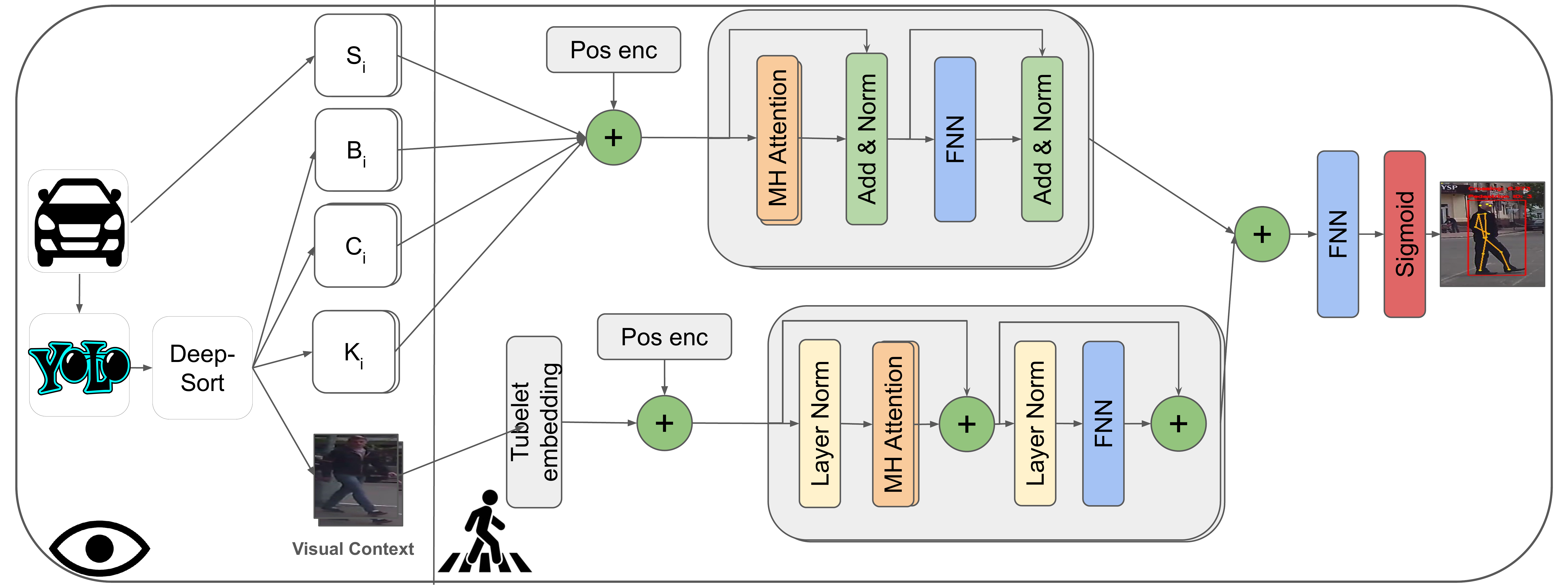}
\caption{Overview on the VIT-Ped Architecture}
\label{fig:full_system}
\end{figure*}

\section*{Methodology}\label{sec3}
\subsection*{Problem Formulation} 
The pedestrian intention prediction task is defined by: given a sequence of data about the last n frames, we need to predict and estimate how likely a pedestrian will cross the road to help ego-vehicles take better decisions. Our goal is to predict the probability of crossing $C_i \in \{0, 1\}$ using the past n time stamps. 

 In our different models we use different features to help our models to predict $C_i$ for each pedestrian. The features that our model use are:
    \begin{itemize}
      \item \textbf{Bounding box}: which is denoted as the top-left and bottom-right coordinates for pedestrian \textit{i}:
        $$B_i = \{b_{i}^{t-m},b_{i}^{t-m+1},...,b_{i}^{t}\}$$
      \item \textbf{Center}: which is the center of the bounding box of pedestrian \textit{i}:
        $$C_i = \{c_{i}^{t-m},c_{i}^{t-m+1},...,c_{i}^{t}\}$$
      \item \textbf{Pose keypoints}: which represent 18 keypoints in pedestrian \textit{i}
        $$K_i = \{k_{i}^{t-m},k_{i}^{t-m+1},...,k_{i}^{t}\}$$
      \item \textbf{Speed of ego-vehicle}:
        $$S_i = \{s^{t-m},s^{t-m+1},...,s^{t}\}$$
      \item \textbf{Local context}: A cropped version of the frame that shows what is around pedestrian i
        $$L_i = \{l_{i}^{t-m},l_{i}^{t-m+1},...,l_{i}^{t}\}$$
      \item \textbf{Global context}: The full frame
        $$G_i = \{g^{t-m},g^{t-m+1},...,g^{t}\}$$
      \item \textbf{Local surround}: A cropped version of the frame where we mask pedestrian \textit{i} so his pixels are all of the same color
        $$A_i = \{a_{i}^{t-m},a_{i}^{t-m+1},...,a_{i}^{t}\}$$
    \end{itemize}
    Each input type has a sequence of length m+1 and t is any time in the range of 1-2 seconds before the action of crossing or not crossing.
\subsection*{Data Preparation and Preprocessing}
    In our experiments we used Joint attention in autonomous driving (JAAD) \cite{kotseruba2016joint} dataset to acquire the features mentioned above. JAAD dataset consists of 346 short video clips (5-10 sec long) extracted from over 240 hours of driving footage filmed in North America and Eastern Europe.The videos were recorded with an on-board camera at 30 FPS and they captured the behaviour of ~1\textit{k} different pedestrians.
    
    Different input types require different methods of preprocessing. 
\subsubsection*{Bounding Boxes \& Center} These features provide us with the information of where the pedestrian resides currently in a frame and tell us more information regarding his/her movement across sequential frames. These features are normalized by a technique called delta encoding in which the bounding boxes and centers from the first frame are subtracted from the last n-1 frames and then those bounding boxes B and centers C in the first frame are removed. Each bounding box is a vector of 2D coordinates which contains the top left and bottom right coordinates, i.e.,
        $$b_i^{t-m} = \{x_{itl}^{t-m},y_{itl}^{t-m},x_{ibr}^{t-m},y_{ibr}^{t-m}\}$$
        where $x_{itl}, y_{itl}$ denotes the top-left point and $x_{ibr}, y_{ibr}$ denotes the bottom-right point.
        
        Each center is a vector of a single 2D coordinate, i.e.,
        $$c_i^{t-m} = \{x_{i}^{t-m},y_{i}^{t-m}\}$$
\subsubsection*{Pose Keypoints} The pose keypoints represent the target pedesrian's posture and detailed motion at each frame. However, the ground truth pose keypoints were not provided by JAAD dataset but it was provided by the PedestrianActionBenchmark repo mentioned in \cite{kotseruba2021benchmark}. A single pose keypoint vector k is a 36D vector of 2D coordinates which contains 18 pose joints, i.e., 
        $$k_i^{t-m} = \{x_{i1}^{t-m},y_{i1}^{t-m},...,x_{i18}^{t-m},y_{i18}^{t-m}\}$$

\subsubsection*{Speed of Ego-Vehicle} The speed of an ego-vehicle S is provided by JAAD, however; it is categorized into 5 categories which are stopped, moving\_slow, moving\_fast, decelerating and accelerating.

\subsubsection*{Local Context} Local Context $L_{ij}$ provides visual features regarding the target pedestrian and some of surroundings around him/her if the enlargement ratio of the bounding box of pedestrian \textit{i} is greater than 1. $L_{ij}$ consists of a sequence of RGB images of different sizes which are ([112,112],[128,128],[224,224]) pixels around the target pedestrian. The images are normalized by dividing by 255.

\subsubsection*{Local Surround} Local Surround $A_{ij}$ is similar to local context however the difference is that the target pedestrian is masked out by greying out the bounding box related to that pedestrian, however; the surroundings remain intact. The images are normalized by dividing by 255.

\subsubsection*{Global Context} Global Context $G_{ij}$ represents the entire frame which was resized into one of the following sizes ([112,112],[224,224]) pixels. The images are normalized by dividing by 255. The objective of that input is that it offers the visual features required to show the different interactions between different road users and the road.
        
\subsection*{Perception} 
    Before diving into our model architecture, we prepared our testing environment. We used pre-recorded videos of different scenarios in which people cross/not cross the roads to see if our models predict the intention correctly and whether it was before the event by \(>=\)1 second. Some of the videos were from JAAD dataset and we also recorded some videos by ourselves in our GUC campus. Therefore, we used YOLOv8 from ultralytics to detect pedestrians, estimate their pose keypoints and track them by giving each pedestrian a unique id. Then, we prepared the inputs for the model being tested and fed them to it and tested whether the model was able to predict intentions correctly or not. 
    
\subsection*{Model Architecture} 
    There are two modalities of data that are used in our models. For non visual data like vehicle speed, bounding box, center, keypoints coordinates we use a transformer \cite{vaswani2017attention}. For visual based data like the local context, global context and local surround we use the video vision transformer (ViViT)\cite{arnab2021vivit} to work with such data. The different features resulting from the different transfromers mentioned above are fused in different ways. The modules that we have are:

\subsubsection*{Transformer Module}
We use the vanilla transformer encoder \cite{vaswani2017attention} for all non visual data. First, we concatenate the non visual data across their last dimension and then apply positional encoding \eqref{eq:sigmoid} to them (the one used in \cite{vaswani2017attention}) to maintain the order of sequences of non visual data across sequential frames so the input is a 3D array of shape (batch size, number of frames, features). Then, we feed the encoded data to the transformer encoder which contains N encoder layers so that the self-attention mechanism allows the model to weigh the importance of different input sequence parts and capture different relationships between them.
        \begin{equation}
        P = \begin{cases}
        \sin\left(\frac{\text{pos}}{10000^{2i/d}}\right), & \text{if } i \mod 2 = 0 \\
        \cos\left(\frac{\text{pos}}{10000^{2i/d}}\right), & \text{if } i \mod 2 = 1
        \end{cases}
        \label{eq:pos_enc}
        \end{equation}

\subsubsection*{ViViT Module}
For visual input, we first perform a special type of embedding called Tubelet Embedding \cite{arnab2021vivit} on the incoming data to capture the information about how a certain patch changes in a video sample over time which is an extension to the patch-based embedding in \cite{dosovitskiy2020image}. The input data, which is a sequence of RGB images, is collected as a 5D array of shape (batch size, number of frames, image height, image width, channels) Then, positional encoding is applied to the embeddings and the result is fed to a ViViT \cite{arnab2021vivit} model where each visual input type (local surround, local context and global context) has a separate ViViT \cite{arnab2021vivit} model.

\subsubsection*{Fusion Module} After feeding the data to the aforementioned modules, their outputs need to be fused together in some way in order to be able to output the probability of whether the pedestrian is about to cross or not. Several ways of fusing data was explored, some of them are concatenating the outputs along the feature dimension then feeding the result to a feed forward neural network or to a Global Average Pooling layer or to an another vanilla transformer encoder or to a LSTM to learn more temporal features regarding the data coming in. Finally, we feed the outputs coming out of the fusion module to a sigmoid \eqref{eq:sigmoid} layer to output the pedestrian's intention probability of whether he/she is about to cross or not.
        \begin{equation}
            \sigma(x) = \frac{1}{1 + e^{-x}}
        \label{eq:sigmoid}
        \end{equation}

        where x represents the outputs from the fusion module.

%% file: Sections/02_Results.tex
\section*{Experiments}
\input{Files/table1}

\input{Files/table2}
\subsection*{Dataset}
Our models were trained and evaluated on the $\text{JAAD}_{all}$ and $\text{JAAD}_{beh}$ , which are 2 subsets of the JAAD dataset \cite{kotseruba2016joint}. It is composed of short driving ego-vehicle clips with annotations like bounding boxes and vehicle annotations. In our work, we added pedestrian keypoints where it captures the location of 18 keypoints' locations throughout the pedestrians' body. Our models were trained to accurately predict pedestrians' crossing intentions with a window of 1-2 seconds prior to the actual crossing action. The training set that was introduced in \cite{kotseruba2016joint} \cite{kotseruba2021benchmark} were used.

\subsection*{Training}
Our small model is trained with an initial learning rate of $3e^{-4}$, a batch size of 32 using Adam optimizer for 100 epochs. With a patience of 5 epochs, the learning rate was reduced by a factor of 0.2 if the learning rate plateaued. Also with a patience of 15 epochs, we used an early stopping callback to stop as soon as our model starts to overfit. To counter the imbalanced data in the JAAD dataset we used class weights for our loss function which is the binary crossentropy function \eqref{eq:binary_crossentropy}.
Our larger model (Ours 3 in table \ref{tab:benchmark}) was trained on both datasets with an initial learning rate of $5e^{-5}$, a batch size of 8 using Adam optimizer for 20 epochs and we also used class weights to reduce the effect of the imbalanced datasets. Then, we fine-tuned the model which was trained on JAAD$_{beh}$ for 20 more epochs, but with no class weights being used; however, we used the ReduceLROnPlateau callback with patience of 5 epochs after which the learning rate was reduced by factor of 0.1 if validation loss did not improve.
\begin{equation}
L(y, \hat{y}) = -\frac{1}{N} \sum_{i=1}^{N} [y_i \log(\hat{y_i}) + (1 - y_i) \log(1 - \hat{y_i})]
\label{eq:binary_crossentropy}
\end{equation}

\subsection*{Metrics}
For our task (pedestrian intention prediction), we use common classification metrics like: Accuracy, Area under the Curve (AUC), F1-score, Precision and Recall.

\subsection*{Models}
Using the repository proposed by \cite{kotseruba2021benchmark} we compare our models to SOTA models for pedestrian intention predictions. Our models are being compared to: ConvLSTM \cite{shi2015convolutional}, CAPformer \cite{lorenzo2021capformer}, PCPA \cite{kotseruba2021benchmark} \& C3D \cite{tran2014learning}.


\subsection*{Ablation Study} 
As indicated in table \ref{tab:ablation}, an ablation study was conducted to compare different fusion strategies as well as trying out different ViViT \cite{arnab2021vivit} architectures which are Spatio-temporal Attention and the Factorised Encoder models. Moreover, we tried out different image sizes and different combinations of different types of inputs to see how the models respond to different modalities fused together. We have also tried out the feature tokenizer idea for tabular data (represented in this paper as the non visual data) from the FT-Transformer \cite{gorishniy2021revisiting} and trained some models on JAAD dataset after balancing the negative and positive samples by resampling from the class with lower number of samples. The following models are considered the most important ones from our experiments:
\begin{itemize}
    \item \textbf{Ours 4}: We have used the Factorised Encoder architecture for the local surround and the global context inputs as well as a vanilla transformer for the rest of non visual data (bbox, pose, center) and fused the outputs together using Global Average Pooling.

    \item \textbf{Ours 5}: Instead of training our model on just 30-60 frames preceding the crossing event, we pretrained our model on the entire dataset, allowing the model to predict the passing behavior from as early as 500 frames before it happens. Following this, we experimented with several fine-tuning techniques to enhance the model's accuracy and efficiency on the dataset.
    
    \item \textbf{Ours 6}: We have trained a vanilla transformer encoder of 2 layers and 4 heads using bounding boxes only as inputs.

    \item \textbf{Ours 7}: This model is an ensemble of 3 other models which we pretrained ourselves and then we froze those models and added a sigmoid layer which takes the outputs of those frozen pretrained models and trained this layer to output the correct probabilities. Two of these models were trained on the balanced version of both datasets in which one of those 2 models was a factorised encoder model (Ours 10) and the other was a Spatio-temporal Attention model (Ours 12). However, the third model was trained on the original version of the datasets (Ours 11). 

    \item \textbf{Ours 8}: We used the FT-Transformer for non visual data (bbox, pose, center, speed) to convert  numerical features to embeddings as well as add cls token to the inputs to see how will that affect the outputs.

    \item \textbf{Ours 9}: Rather than allowing later frames attend to earlier ones, we applied attention masking to ensure that each frame can only attend to frames that came before it to preserve the temporal ordering of the sequence.

\end{itemize}

%% file: Files/table1.tex

\begin{table}[ht]
\centering
\begin{tabularx}{\textwidth}{@{}l*{10}{S[table-format=2.3]}@{}}
\toprule
\multirow{2}{*}{Model} & \multicolumn{5}{c}{JAAD All} & \multicolumn{5}{c}{JAAD Beh} \\
\cmidrule(lr){2-6} \cmidrule(lr){7-11}
 & {Acc} & {AUC} & {F1} & {Precision} & {Recall} & {Acc} & {AUC} & {F1} & {Precision} & {Recall} \\
\midrule
ConvLSTM & 0.585 & 0.582 & 0.327 & 0.228 & 0.578 & 0.590 & 0.530 & 0.701 & 0.644 & 0.767 \\
CAPformer & 0.731 & 0.778 & 0.525 & 0.380 & \textbf{0.85} & 0.511 & 0.544 & 0.516 & \textbf{0.68} & 0.415 \\
PCPA & 0.761 & 0.784 & 0.545 & 0.408 & 0.822 & 0.536 & 0.490 & 0.646 & 0.619 & 0.675 \\
C3D & 0.854 & 0.764 & 0.600 & 0.574 & 0.628 & 0.584 & 0.477 & 0.732 & 0.614 & 0.907 \\
\midrule 
Ours 1& 0.81 & 0.74 & 0.51 & 0.47 & 0.55 & 0.62 & \textbf{0.61} & \textbf{0.77} & 0.63 & \textbf{0.99}\\
Ours 2 (non-visual)& 0.84 & \textbf{0.81} & 0.51 & 0.52 & 0.5 & \textbf{0.64} & 0.58 & \textbf{0.77} & 0.65 & 0.93\\ 
Ours 3 & \textbf{0.86} & 0.771870084 & \textbf{0.61} & \textbf{0.58} & 0.644010195 & 0.6198830409356725 & 0.5506982795242141 & 0.7311019180142911 & 0.66 & 0.8258283772302464\\

\bottomrule
\end{tabularx}
\caption{\label{tab:benchmark}Comparison of Various Models on JAAD Dataset Metrics}
\end{table}

%% file: Files/table2.tex
\begin{table}[ht]
\centering
\begin{tabularx}{\textwidth}{@{}l*{10}{S[table-format=2.3]}@{}}
\toprule
\multirow{2}{*}{Model} & \multicolumn{5}{c}{JAAD All} & \multicolumn{5}{c}{JAAD Beh} \\
\cmidrule(lr){2-6} \cmidrule(lr){7-11}
 & {Acc} & {AUC} & {F1} & {Precision} & {Recall} & {Acc} & {AUC} & {F1} & {Precision} & {Recall} \\
\midrule
Ours 4 (factorised encoder)& 0.820707071 & 0.790919185 & 0.592367443 & 0.491591928 & 0.745114698 & 0.59702286 & 0.527008549 & 0.714393369 & 0.641841571 & 0.805437553\\ 
Ours 5 (pretrained) & 0.83 & 0.85 & 0.56 & 0.5 & 0.63 & 0.61 & 0.64 & 0.67 & 0.61 & 0.63\\
Ours 6 (bboxes only) & 0.779560309 & 0.727819511 & 0.506976744 & 0.416257501 & 0.648258284 & 0.630515683 & 0.564902825 & 0.736642668 & 0.664842681 & 0.825828377 \\ 
Ours 7 (ensemble) & 0.752822341 & 0.786278254	& 0.542354235 & 0.400976007	& 0.837723025	& 0.642743222 & 0.582665144 & 0.742133538 & 0.676696991	& 0.821580289\\
Ours 8 (ft \& non visual) & 0.745840761 & 0.698013072 & 0.462118831 & 0.366766467 & 0.624468989 & 0.532163743	& 0.53569058 & 0.582542694 & 0.659505908 & 0.521665251 \\ 
Ours 9 (attention masking) & 0.81 & 0.74 & 0.51 & 0.47 & 0.55 & 0.62 & 0.61 & 0.77 & 0.63 & 0.991 \\ 
Ours 10 & 0.816102198 & 0.789468106 & 0.587333333 & 0.483269336 & 0.748513169 & 0.514088251 & 0.512113689 & 0.572497661 & 0.636836629 & 0.519966015 \\
Ours 11  & 0.840463458 & 0.783806792 & 0.604274134 & 0.533506831 & 0.696686491	& 0.557150452 & 0.570212935 & 0.594252314 & 0.696347032 & 0.51826678 \\
Ours 12 & 0.820261438 & 0.78763596 & 0.589273591 & 0.490672696 & 0.737468139	& 0.518341308 & 0.482403621 & 0.619007569 & 0.612822648 & 0.625318607 \\
\bottomrule
\end{tabularx}
\caption{\label{tab:ablation}Ablation Study}
\end{table}

%% file: Sections/03_Discussion.tex
\begin{figure*}[ht]
\centering
\includegraphics[width=\textwidth]{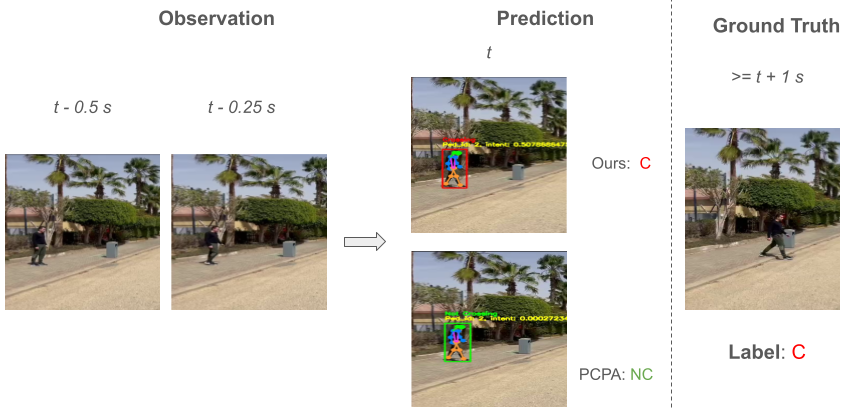}
\caption{Our model predicts that the intention of the pedestrian correctly unlike PCPA. Red bounding box means the pedestrian is about to cross and green means not crossing.}
\label{fig:examples}
\end{figure*}
\begin{figure*}[ht]
\centering
\includegraphics[width=\textwidth]{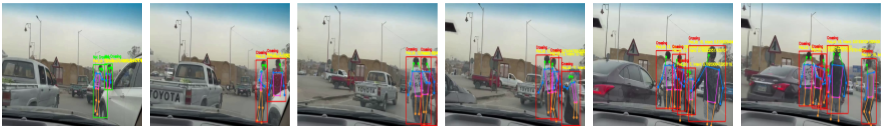}
\caption{Chaotic Environment. The frames are separated from one another by a window of 1-2 seconds. Red bounding box means the pedestrian is about to cross/crossing and green means not crossing.}
\label{fig:chaotic_env}

\end{figure*}

\section{Results}\label{sec4}

\subsection*{Quantitative Results} 
    Table \ref{tab:benchmark} show the quantitative results on JAAD$_{all}$ and JAAD$_{beh}$ datasets, respectively. These tables compare our proposed model to the benchmark models: ConvLSTM \cite{shi2015convolutional}, CAPformer \cite{lorenzo2021capformer}, PCPA \cite{kotseruba2021benchmark} \& C3D \cite{tran2014learning}. The proposed models achieved the best scores in accuracy, auc and f1 score in both datasets and these metrics are the most necessary ones for  binary classification.

    Table \ref{tab:ablation} show the quantitative results of the ablation studies on JAAD$_{all}$ and JAAD$_{beh}$ datasets, respectively.
    
\subsection*{Qualitative Results} 
    Figure \ref{fig:examples} provides the qualitative results in which we compare our model to PCPA using some pre-recorded videos of real-life scenarios. In the provided examples in figure \ref{fig:examples}, our model predicts correctly the ground truth intentions unlike PCPA. By taking closer look at the examples, we hypothesize that the reason for that is that our models were trained on local surround and global context inputs; therfore, the models are more aware of the surroundings regarding the target pedestrian as well as more robust to whether the target pedestrian is looking towards the ego-vehicle or not due to the fact that the target pedestrian is grayed out in the local surround input. The detailed results of the experiment can be viewed from the media \url{https://youtu.be/CssklQmzP6c?si=93LcZ9fbboFE4i3Q}.
    
    Figure \ref{fig:chaotic_env} shows an example of the output of our ensemble model (Ours 7) in a chaotic environment where the pedestrians are already on the road and the cars are everywhere. The frames are separated from one another by a window of 1-2 seconds. This scenario is not similar to almost all of the scenarios in JAAD dataset on which the model was trained which shows that our model can be applicable in real-life situations. The detailed results of the experiment can be viewed from the media \url{https://youtu.be/FKbzKw1TSDc?si=60jQNtF-G2zaxNce}.